\documentclass[a4paper]{article}

\usepackage{INTERSPEECH2022}
\usepackage{graphicx}
\usepackage{tabularx}
\usepackage{soul}
\usepackage{tikz}
\usepackage{amsmath}
\usepackage{xspace}
\usepackage[normalem]{ulem}
\usepackage{todonotes}

\usepackage{epstopdf}
\usepackage[utf8]{inputenc}

\usepackage{hyperref}
\usepackage{xstring}
\usepackage{hyperref}
\usepackage{xcolor}
\usepackage{multirow}

\usepackage{todonotes}

\title{End-to-end model for named entity recognition from speech without paired training data}
\name{Salima Mdhaffar, Jarod Duret, Titouan Parcollet, Yannick Estève}
\address{
  LIA - Avignon Université, France}
\email{\{firstname.lastname\}@univ-avignon.fr}

\begin{document}

\maketitle
\begin{abstract}
Recent works showed that end-to-end neural approaches tend to become very popular for spoken language understanding (SLU). 
Through the term end-to-end, one considers the use of a single model optimized to extract semantic information directly from the speech signal.
A major issue for such models is the lack of paired audio and textual data with semantic annotation.
In this paper, we propose an approach to build an end-to-end neural model to extract semantic information in a scenario in which zero paired audio data is available.
Our approach is based on the use of an external model trained to generate a sequence of vectorial representations from text. 
These representations mimic the hidden representations that could be generated inside an end-to-end automatic speech recognition (ASR) model by processing a speech signal.
A SLU neural module is then trained using these representations as input and the annotated text as output.
Last, the SLU module replaces the top layers of the ASR model to achieve the construction of the end-to-end model.
Our experiments on named entity recognition, carried out on the QUAERO corpus, show that this approach is very promising, getting better results than a comparable cascade approach or than the use of synthetic voices.

\end{abstract}
\section{Introduction}
Spoken language understanding (SLU) is a very popular topic related to artificial intelligence (AI) widely used in various real-life applications e.g., personal assistants or smart devices.
Since 2018, end-to-end neural approaches have been introduced to approach these tasks~\cite{serdyuk2018towards,ghannay2018end,haghani2018audio,lugosch2019speech,desot2019towards}. 
While classical SLU systems are built on a cascade approach composed of at least two modules e.g., an automatic speech recognition system and another one for natural language understanding, end-to-end architectures, on the other hand, consider the use of a single neural model optimized to extract semantic information directly from the speech signal.

The two main purposes of the introduction of end-to-end approaches were (i) the joint optimization of the automatic speech recognition (ASR) and the natural language understanding (NLU) parts to the final task, and (ii) the limitation of the error propagation due to the bottleneck that constitutes the ASR output. Nevertheless, such end-to-end SLU models need to be trained on large amounts of data, while a major issue is the lack of paired audio and textual data with semantic annotation: producing a large amount of annotated speech data remains difficult and expensive. 

However, for some tasks including named entity recognition (NER), large amounts of annotated data may be available in the textual format. Hence, with this paper, we propose to leverage the latter textual-only data to conceive an end-to-end neural model extracting semantic information from speech in a named entity recognition scenario in which no paired textual-audio data is accessible.

\section{Related work}
\label{sec:relatedwork}
The lack of resource is recurrent for SLU tasks. 
For end-to-end approaches this gap is even more critical than for cascaded ones due to the bi-modal nature of the necessarily paired training data.
Several solutions have been recently proposed in the literature to overcome this issue. 
In~\cite{caubriere2019curriculum}, the authors propose to apply a curriculum-based transfer learning approach to take benefit of existing semantic annotations related to more generic SLU tasks than the targeted one.
In~\cite{lugosch2020,huang2020leveraging} the authors suggest to generate synthetic speech in order to extend a small training dataset of paired data of audio and semantic annotation. This approach has also been proposed for data augmentation in ASR~\cite{laptev2020you} or with speech translation~\cite{kano2020end}

Other approaches aim to merge acoustic and text embeddings. In~\cite{huang2020leveraging}, acoustic embeddings for intent classification are tied to fine-tuned BERT text embeddings, while in~\cite{jia2020large} a multi-task architecture is designed to map the acoustic feature sequence into a semantic space shared for both speech recognition and SLU.

More recently, \cite{pasad2021use} presented a study that explores several ways intended to inject external speech and/or text unlabeled data for the task, to train cascaded or end-to-end named entity recognition systems. 
Their experiments showed that, with access to unlabeled speech or transcribed speech, end-to-end models outperform pipeline models.

To the best of our knowledge, our work is the first successful attempt at conceiving an end-to-end model for named entity recognition from speech without the use of paired training data\footnote{This paper has been submitted to INTERSPEECH 2022}.

\section{Proposed approach}
\label{sec:proposedapproach}

The approach presented below is dedicated to SLU for which no paired data (i.e., audio with transcriptions annotated with semantic information) is available.
In this study, we focus on NER from speech, but we assume that this approach could be extended to other similar SLU tasks. 
More precisely, NER from speech consists of recognizing words from an input speech source, detecting word sequences that support a named entity, and categorizing this entity.

In our low resource SLU scenario, an end-to-end model for ASR and a corpus of textual documents with named entity annotations but without the corresponding audios are available.

Our approach is based on the use of an external model trained to generate a sequence of vectorial representations from text. 
These representations mimic the hidden representations that could be generated inside an end-to-end automatic speech recognition model by processing a speech signal.
A SLU neural module is then trained to use these representations as input and the annotated text as output.
Last, the SLU module replaces the top layers of the ASR model to achieve the construction of the end-to-end model.

To generate the simulated ASR hidden representations (or ASR embeddings), we train a sequence-to-sequence neural model, called \textsl{Text-to-ASR-Embeddings} model.
Such an approach can be compared to some recent propositions mentioned in section~\ref{sec:relatedwork} that use synthetic voices to feed an ASR end-to-end model.
We motivated our proposition for different reasons.
First, the use of synthetic speech introduces some artifacts in the input of the ASR model. 
If the ASR model is fine-tuned on such synthetic voices, these artifacts will degrade the capability of the model to process natural voices.
A solution to avoid this consists of freezing the weights of the bottom layers and only update the weights of the higher layers, in which the semantic is better encoded.
Since the bottom layers were optimized to process natural speech, the quality of the embeddings computed from synthetic speech is not guaranteed, and can introduce a gap between embedding computed from natural and computed from synthetic speech.
With our approach, we aim to reduce this gap.
In addition, our approach needs less computation at training time than the ones based on synthetic speech, since we avoid the use of a consequent number of lower layers.

\begin{figure}[htb!]
\centering

\includegraphics[width=\columnwidth]{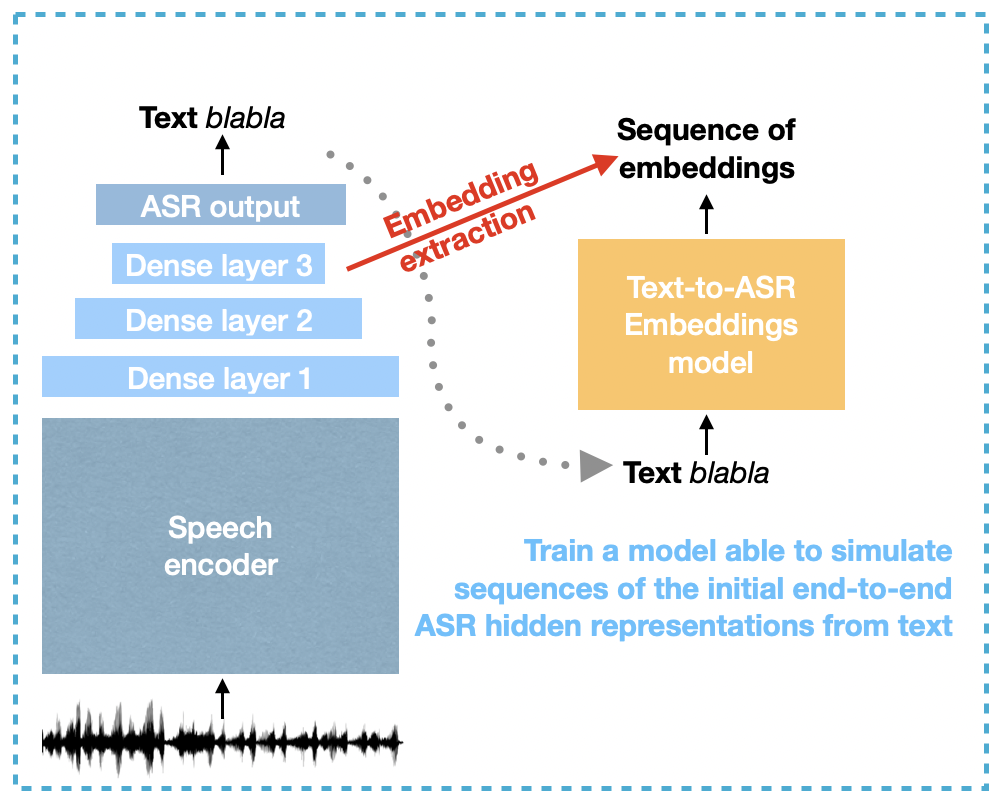}
\caption{ASR embeddings extraction to train the \textsl{Text-to-ASR-Embeddings} model} 
\label{fig:1_trainingTTsE}
\end{figure}

To train this neural network, we must produce a training dataset composed of pairs of transcriptions, used as input, and sequences of ASR embeddings, used as output. 
To produce this training dataset, the end-to-end ASR model is used to transcribe its training dataset.
For each transcribed utterance, we extract a sequence of ASR embeddings from a hidden layer, and associate this ASR embedding sequence to the automatic transcription.
When the entire ASR training data has been processed, the ASR embedding sequences and their associated automatic transcriptions are used to train the \textsl{Text-to-ASR-Embeddings} model, as illustrated in Figure~\ref{fig:1_trainingTTsE}

\begin{figure}[htb!]
\centering
\includegraphics[width=\columnwidth]{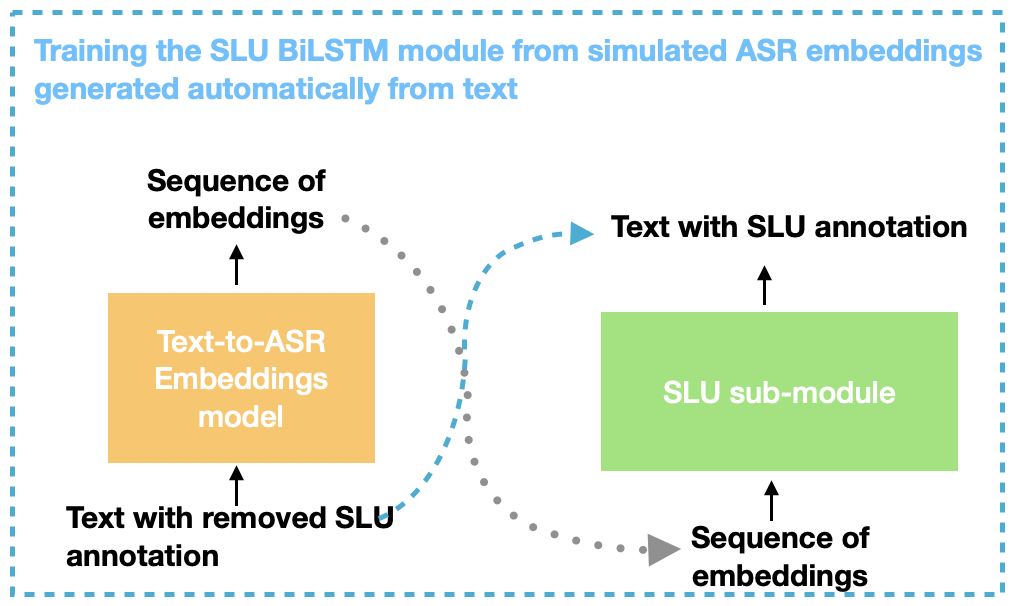}
\caption{Training the SLU sub-module from simulated ASR embeddings} 
\label{fig:2_trainingNERmodule}
\end{figure}

At this stage, we obtain a module able to simulate ASR embeddings from text.
Our objective is then to train a neural SLU sub-module able to convert such a sequence of ASR embedding into an automatic transcription with SLU annotation, like annotation of named entities.
For this purpose, we exploit the textual dataset with semantic annotation.
For each sentence in this dataset, we first remove the semantic annotation to keep only the sequence of words.
Thanks to the \textsl{Text-to-ASR-Embeddings} model, we transform this sequence of words to a sequence of ASR embeddings.
We iterate this process for all the annotated sentences in the semantic textual dataset.
We get a set of pairs composed of a sequence of ASR embeddings and the corresponding text sequence of words semantically annotated.
Once the entire textual dataset has been processed, we use this data to train an SLU sub-module able to generate a sequence of words semantically annotated from a sequence of ASR embeddings.
This step is illustrated in Figure \ref{fig:2_trainingNERmodule}.

\begin{figure}[htb!]
\centering
\includegraphics[width=0.65\columnwidth]{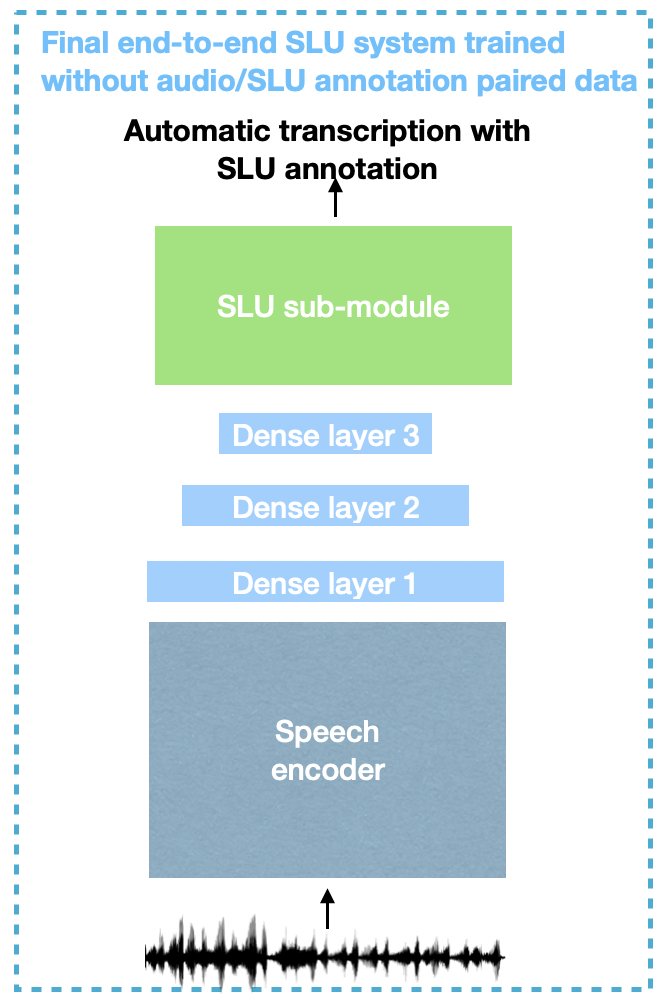}
\caption{Final end-to-end SLU system trained without audio/SLU annotation paired data} 
\label{fig:3_final_e2eNERmodule}
\end{figure}

Finally, we plug the end-to-end ASR and the SLU sub-module, as illustrated in Figure~\ref{fig:3_final_e2eNERmodule}.
In order to merge the ASR model with the SLU sub-module, we keep all the ASR hidden layers needed to generate the ASR embeddings that can be mimicked by the \textsl{Text-to-ASR-Embeddings} model. 
The mimicked hidden layer is then connected to the SLU sub-module.
In the example of Figure~\ref{fig:3_final_e2eNERmodule}, only the ASR output layer is removed.

The final model is an end-to-end model able to transcribe and extract semantic information from speech, while no real paired training data exists.
In the next section, we show how we implemented this approach to recognize named entities from speech.

\section{Experiments}
\label{sec:experiment}

Our experiments were carried out on the French QUAERO data, officially used in the framework of the ETAPE evaluation campaign~\cite{gravier2012etape} and in different research studies \cite{caubriere-etal-2020-named,jannet2015evaluate} dedicated to named entity recognition from speech.

\subsection{Data}

Several publicly available corpora have been used for these experiments.
\subsubsection{ASR training data}
\label{sec:ASR_data}
The ASR training data is composed of different corpora in French language, all collected from TV or radio stations:
\begin{itemize}
    \item REPERE\footnote{The REPERE corpus can be obtained from ELRA under the reference ELRA-E0044} \cite{giraudel2012repere}: The REPERE dataset was created by the organizers of the REPERE evaluation campaign. It is composed of articles from videos (news, political debate and talk-shows) from 7 French TV programs (2 programs from BFMTV channel and 5 from LCP channel) broadcasted between 2011 and 2012.
    \item ESTER2\footnote{The ESTER2 dataset can be obtained from ELRA under the reference ELRA-S0241} : The ESTER2 dataset was created for the ESTER~2 evaluation campaign~\cite{galliano2009ester}. It consists on radio broadcast news, from French, Moroccan and African radio stations.
    \item EPAC\footnote{The EPAC corpus can be obtained from ELDA under the reference ELDA-S0305} \cite{esteve2010epac}:  this data is composed of audio recordings of French radio and TV shows, mainly debates or interviews.

\end{itemize}
The total duration of all these datasets is 197 hours of speech (35.5 hours of REPERE, 91 hours of ESTER2, 70.5 hours of EPAC).

\subsubsection{Text-to-ASR-Embeddings data}

The ASR embeddings used to train the \textsl{Text-to-ASR-Embeddings} model have been extracted from the ASR model by processing its own training data, described in the previous subsection.
The \textsl{Text-to-ASR-Embeddings} model is trained from these ASR embeddings and their corresponding automatic transcriptions computed by the ASR model during the extraction.
The word error rate of the automatic transcriptions got on the 197 hours of the ASR training data is 7.4\%.
No filtering was applied on these transcriptions before training the \textsl{Text-to-ASR-Embeddings} model, but 10\% of the data was removed from the training data to be used as validation data to control the training process of this model.

\subsubsection{Dataset for named entity recognition}
\label{sec:NER_data}
To train and evaluate our named entity recognition systems, we used the QUAERO\footnote{The QUAERO dataset can be obtained from ELRA under the reference ELRA-S0349} dataset \cite{grouin2011proposal}. 

This dataset contains broadcast news speech with text transcription and named entity annotation.
Manual annotation of named entities is available according to eight types: \textit{person}, \textit{location}, \textit{organization}, \textit{product}, \textit{time}, \textit{amount}, \textit{function} and \textit{event}.
In our scenario, we assume that we do not have a speech dataset. 
We use only the textual transcriptions and the named entity annotation in the text, except for a contrastive experiment. 
We divided the QUAERO corpus into three parts (the official distribution contains only two parts): a training corpus, a validation corpus, and a test corpus.
The test corpus is composed of the ten most recent hours, in order to avoid overlaps with the training data.
To ensure the reproducibility of our experimental results, we share the details of our file distribution\footnote{https://github.com/mdhaffar/Named-Entity-Recognition}.
Some statistics about this data are described in Table \ref{tab:stat_QUAERO}.
Figure \ref{fig:example_QUAERO} shows an example of QUAERO dataset.

\begin{table}[h!]
\centering
\begin{tabular}{|l|l|l|l|l|l|}
\hline
      & \#words & \#vocab & \#entities & \#sentence\\ \hline \hline
Train &  993,746       &    33,583     &    89,609   &  90,378 \\ \hline
Dev  &  81,531      &     8,713    &     5,538   & 10,982 \\ \hline
Test   &  119,361     &   13,447      &    10,883    & 4,023 \\ \hline

\end{tabular}
\caption{QUAERO dataset}
\label{tab:stat_QUAERO}
\end{table}

\vspace{-1cm}
\definecolor{color_concept_a}{RGB}{2, 62, 138}    
\definecolor{color_concept_b}{RGB}{230, 57, 70}   
\definecolor{color_concept_c}{RGB}{130, 173, 103} 
\definecolor{color_concept_d}{RGB}{0, 127, 127}   
\definecolor{color_concept_e}{RGB}{247, 37, 133}  
\definecolor{color_concept_f}{RGB}{247, 127, 0}   
\definecolor{color_concept_g}{RGB}{255, 183, 0}  

\newcommand{\openconcept}[2]{\mbox{\textcolor{#1}{
    \textbf{\small{\textless}}\hspace{-4pt}
    \textbf{#2}
}}}
\newcommand{\closeconcept}[1]{\textcolor{#1}{
    ~\textbf{\small{\textgreater}}\hspace{-4pt}
}}

\newcommand{\concept}[3]{\openconcept{#1}{#2}#3\closeconcept{#1}}
\begin{figure}[!ht]
    \centering
    \begin{tikzpicture}
        \node[] at (0, 0) {(a)} ;
        \node[] at (0, -0.6cm) {(b)} ;
        \node[anchor=west,text width=\linewidth - 0.5cm] at (0.5cm, 0) {demain rfi présente le huitième festival de jazz de saint louis au sénégal};
        \node[anchor=north west,text width=\linewidth - 0.5cm] at (0.5cm, -0.3cm) {
        \concept{color_concept_e}{time}{demain}
        \concept{color_concept_d}{organisation}{rfi}
        présente le huitième festival de jazz de
        \concept{color_concept_f}{location}{saint louis} au \concept{color_concept_f}{location}{sénégal}};
    \end{tikzpicture}
    \caption{An example of a QUAERO dataset sample in French of the English sentence \textit{`tomorrow rfi presents the eighth saint louis jazz festival in senegal'}. (a) corresponds to the transcribed sentence. (b) the same sample with its additional named entity tags. Here, `\openconcept{color_concept_b}{time}' is an opening tag starting the support word sequence `demain' (tomorrow in English) and expressing that this word sequence is associated with the \textsl{time} entity. The character \textit{`$>$'} represents the closing tag and it is used to close all entity tags.}
    \label{fig:example_QUAERO}
\end{figure}

\subsection{Model implementation}
In this section, the different models used in our experiments are described.
All the models were implemented thanks to the SpeechBrain toolkit \cite{parcollet2022speechbrain}. 
For reproducibility purpose, we will also make the recipe available \footnote{note for reviewers: the recipe will be released before the Interspeech conference}.

\subsubsection{End-to-end ASR architecture}
\label{sssec:e2eASRarchi}
The ASR used in this paper is composed of a large pre-trained French wav2vec~2.0 model, with three additional hidden layers of 1024, 512 and 80 units, associated to Leaky ReLU activation functions, and last a softmax output layer.
wav2vec~2.0 is a model pre-trained through self-supervision~\cite{baevski2020wav2vec}. 
The loss function used at each fine-tuning step is the Connectionist Temporal Classification (CTC) loss function \cite{graves2006connectionist}.
This ASR model is a state-of-the-art ASR model for French: the pretrained wav2vec~2.0 model is the LeBenchmark-7K model presented in~\cite{evain2021lebenchmark} and publicly accessible~\footnote{https://huggingface.co/LeBenchmark}.

\subsubsection{Text-to-ASR Embeddings model architecture}

In this study, the model used to mimic the ASR hidden representations is based on the well-known Tacotron2 neural architecture~\cite{tacotron2} initially designed for speech synthesis, also called text-to-speech (TTS) task.
Tacotron2 is composed of a first module that maps character embeddings to mel-scale spectrogram, followed by a second module, a vocoder that transforms the mel-scale spectrogram into audio waveforms.
To generate the simulated ASR hidden representations (or ASR embeddings), we only need this first module: instead of generating a mel-scale spectrogram, the recurrent sequence-to-sequence network that composes this module is trained to generate sequences of ASR embeddings.
The network consists of a stack of 3 convolutional layers, followed by a batch normalization and ReLu activations. 
The goal of these convolutional layers is to model longer-term context in the input character sequence. 
Then the output is fed to a bi-directional LSTM layer to generate the encoded features.

We use text input with basic rule-based text normalization for French language.
For a contrastive experiment presented below, we also trained a Tacotron2 text-to-speech model on the SynPaFlex corpus data~\cite{sini2018synpaflex}. 
This corpus is a French audiobook dataset containing eighty-seven hours of speech, recorded by a single speaker reading audiobooks of different literary genres.

\subsubsection{SLU sub-module}

The SLU sub-module is implemented through a BiLSTM architecture.
This neural architecture learns to transcribe and recognize named entities from the mimicked ASR embeddings generated automatically from text.
The Bidirectional LSTM model is composed of a stack of 5 biLSTM layers, with 512 dimensions each.

\subsection{Experimental results}
This section describes the different results we obtained.
The evaluation of the NER task uses the Named Entity Error Rate (NEER), which is estimated like the classical Word Error Rate (WER) but applied to semantic concepts instead of words. 
Insertions, substitutions, and deletions are all counted as errors. 
All our experiments are evaluated on the test partition defined in Section \ref{sec:NER_data}.

The first row in Table~\ref{tab:results} presents the results of our proposed approach to train an end-to-end SLU model (47.5\% NEER for dev and 39.1\% for test) when the \textsl{Text-to-ASR-Embeddings} model is used to mimic the 80-dimensional embeddings of the last hidden layer presented in section~\ref{sssec:e2eASRarchi}.

We also evaluate the use of synthetic speech to train the end-to-end NER model. 
In a first experiment, all the weights of the end-to-end ASR model are updated during the training process. On the test set, this system gets a NEER of 62.7\%, very higher than the NEER got by our system.
In a second experiment, we freeze all the layers of the ASR speech encoder (i.e. the wav2vec~2.0 part) before starting the NER training. The objective is to limit the possible negative effects of the synthetic speech artifacts to the speech encoder weights: the results are worse, with a NEER of 92.5\%.

Then, we consider the availability of paired audio data.
The SLU sub-module is trained by using the embeddings generated from the ASR model by processing the real audio data. 
These results can be seen as the Oracle NEER we could reach if the \textsl{Text-to-ASR-Embeddings} model was able to produce ideal embeddings.

\begin{table}[h!]
\centering
\begin{tabular}{|l|l|l|}
\hline
Training data   & Dev & Test \\ \hline \hline
ASR embeddings simulation (ours)   &  47.6   &  39.1    \\ \hline
Synthetic speech (all weights are updated) &  65.2   &   62.7   \\ \hline
Synthetic speech (frozen speech encoder)& 86.4 & 92.5\\ \hline 
\hline
Oracle (real audio)               &  45.9   &  34.1    \\ \hline 
\end{tabular}
\caption{Evaluation in NEER (\%) of our approach to train an end-to-end NER model without paired training data compared to other approaches using speech synthesis, and compared to the ideal scenario when paired data is available}
\label{tab:results}
\end{table}

We have shown that our approach makes possible the training of an end-to-end model for named entity recognition from speech without paired training data (using zero audio data).

We would like now to compare its performance to the one got by a cascaded approach trained in the same conditions, since such a approach does not need paired (audio/text) training data.
In this experiment, we use the same neural architecture as the SLU sub-module used on the top of the end-to-end proposed system to train a text-to-text NER model.
We apply this NER model to the automatic transcription of our end-to-end ASR model. 
This ASR model obtains a WER of 12.85\% on the development set and 8.6\% on the test set, showing that the test set seems easier to process than the development set.
We also apply this NER model to the manual transcriptions. 
Table \ref{tab:text_EnER} presents the results of these experiments.
As expected, ASR errors impact the NER model performance, and we can notice that the cascaded approach applied to speech gets worse results than the end-to-end approach trained through the proposed approach.
We are aware that a state-of-the-art cascaded system could strongly take advantage of self-supervised learning-based models like BERT~\cite{devlin2018bert}, but this is out of the scope of this study.

\begin{table}[h!]
\centering
\begin{tabular}{|l|l|l|}
\hline
 & Dev & Test \\ \hline \hline
Text-to-text NER on manual transcripts &  39.6   &  30.0 \\ \hline
\hline
Text-to-text NER on automatic transcripts &  48.0   &    40.2  \\ \hline
\end{tabular}
\caption{Results in term of NEER(\%) got by NER module applied to text.}
\label{tab:text_EnER}
\end{table}

\vspace{-1cm}

\section{Conclusion}
In this paper, we propose an approach to train an end-to-end model for named entity recognition from speech without paired audio and textual data with semantic annotation. 
Our approach, based on artificial ASR embeddings generated from text, exhibits highly promising results outperforming alternative approaches based on the use of synthetic speech.

We consider that this approach can be extended to similar SLU tasks such as slot filling, and opens new perspectives in different use cases where enriching or adapting the linguistic knowledge captured by an end-to-end ASR model is needed.

\section{Acknowledgements}
This work was partially funded by the European Commission through the SELMA European project under grant number 957017.
This work was granted access to the HPC resources of IDRIS under the allocation 2021-AD011012551 made by GENCI.

\bibliographystyle{IEEEtran}

\bibliography{mybib}

\end{document}